\documentclass[10pt,twocolumn,letterpaper]{article}

\usepackage{iccv}
\usepackage{times}
\usepackage{epsfig}
\usepackage{graphicx}
\usepackage{amsmath}
\usepackage{amssymb}
\usepackage{multicol}
\usepackage[switch]{lineno}
\usepackage{color}
\usepackage{multirow}
\usepackage[marginal]{footmisc}
\usepackage{bbding}

\usepackage[pagebackref=true,breaklinks=true,letterpaper=true,colorlinks,bookmarks=false]{hyperref}

\iccvfinalcopy % *** Uncomment this line for the final submission

\def\iccvPaperID{11} % *** Enter the ICCV Paper ID here

\ificcvfinal\fi

\begin{document}

%%%%%%%%% TITLE
\title{Iterative Robust Visual Grounding with Masked Reference based Centerpoint Supervision}

\author{Menghao Li\textsuperscript{1} \textsuperscript{*} \and Chunlei Wang\textsuperscript{1} \thanks{Contribute Equally.} \and Wenquan Feng\textsuperscript{1}\and Shuchang Lyu\textsuperscript{1}\and Guangliang Cheng\textsuperscript{2~\Envelope}\and Xiangtai Li\textsuperscript{3}\and Binghao Liu\textsuperscript{1}\and Qi Zhao\textsuperscript{1~\Envelope}
\and \\
\textsuperscript{1} Beihang University \textsuperscript{2} University of Liverpool \textsuperscript{3} S-Lab, Nanyang Technological University \\
{\tt\small \{sy2102227, wcl\_buaa, buaafwq, lyushuchang, liubinghao, zhaoqi\}@buaa.edu.cn}\and {\tt\small \{Guangliang.Cheng\}@liverpool.ac.uk}\and {\tt\small \{xiangtai.li\}@ntu.edu.sg}
}

\maketitle
%%%%%%%%% ABSTRACT
\begin{abstract}
Visual Grounding (VG) aims at localizing target objects from an image based on given expressions and has made significant progress with the development of detection and vision transformer.
However, existing VG methods tend to generate \textbf{false-alarm} objects when presented with inaccurate or irrelevant descriptions, which commonly occur in practical applications.
Moreover, existing methods fail to capture fine-grained features, accurate localization, and sufficient context comprehension from the whole image and textual descriptions. 
To address both issues, we propose an Iterative Robust Visual Grounding (\textbf{IR-VG}) framework with Masked Reference based Centerpoint Supervision (MRCS).
The framework introduces iterative multi-level vision-language fusion (IMVF) for better alignment.
We use MRCS to ahieve more accurate localization with point-wised feature supervision.
% to comprehensively amplify the vision-language understanding and alignment.
%
Then, to improve the robustness of VG, we also present a multi-stage false-alarm sensitive decoder (MFSD) to prevent the generation of false-alarm objects when presented with inaccurate expressions. 
%
%The proposed framework is evaluated on five {\it \textbf{regular}} VG datasets and two newly constructed {\it \textbf{robust}} VG datasets. 
Extensive experiments demonstrate that IR-VG achieves new state-of-the-art (SOTA) results, with improvements of 25\% and 10\% compared to existing SOTA approaches on the two newly proposed robust VG datasets. Moreover, the proposed framework is also verified effective on five {\it \textbf{regular}} VG datasets. Codes and models will be publicly at \url{https://github.com/cv516Buaa/IR-VG}.
\end{abstract}
\vspace{-0.7cm}

\section{Introduction}
\par
\begin{figure}
\begin{center}
\centering
\includegraphics[scale=0.17]{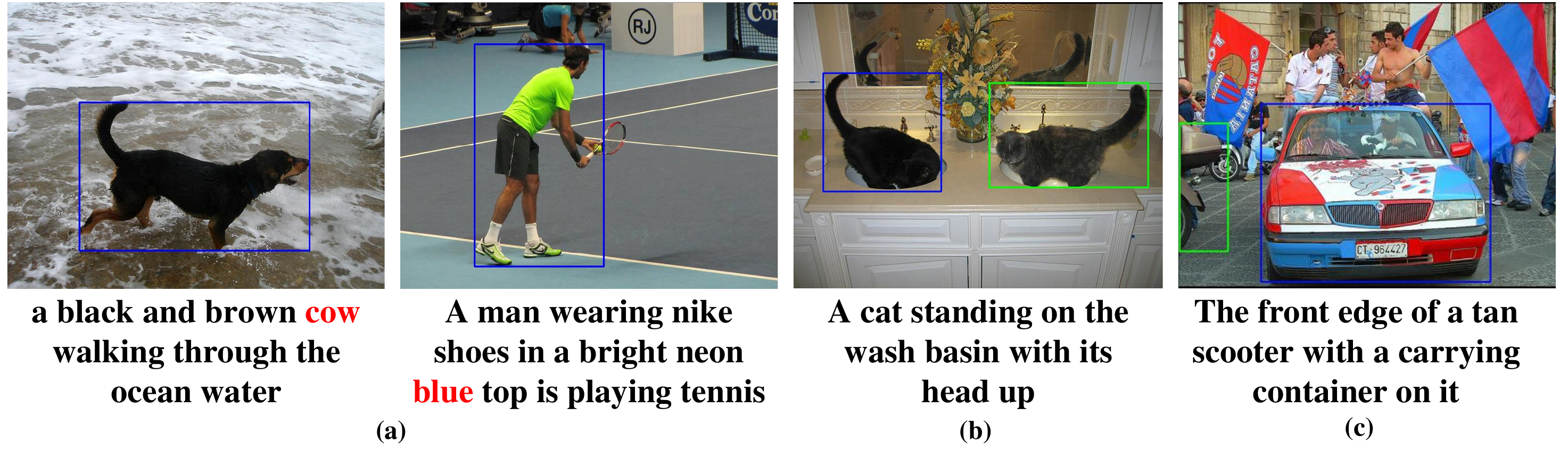}
\end{center}
   \caption{Weaknesses illustration of the existing VG approaches. Green and blue boxes represent groundtruths and prediction. Green boxes and blue boxes represent the ground truths and predictions, respectively. (a) Failure cases occur when an irrelevant or inaccurate description is provided. (b) Fine-grained features are not captured or misunderstood. (c) Predictions are not correlated with the given descriptions.}
\label{fig:failurecases}
\end{figure}

 Visual Grounding (VG) is a crucial computer vision task gaining significant attention due to its potential for enabling practical applications such as robot navigation~\cite{robot_navigation} and visual dialog~\cite{visual_dialog1,visual_dialog2}. VG aims to locate a target object within an image based on the given language reference expressions by incorporating information from both textual and visual modalities. However, existing VG methods suffer from false-alarm issues, where they assume that the referred object always exists in the image, leading to inaccurate or wrong targets being detected when irrelevant or inaccurate textual expressions are provided, shown in Fig.~\ref{fig:failurecases} (a).
\par Previous works~\cite{previous1,previous2,previous3} have made significant progress in VG through various techniques. However, the task of cross-modal learning involved in the VG task remains challenging, and current approaches can be broadly divided into two main categories: two-stage methods~\cite{CMN, NMTree, Ref-NMS,Two-Branch} and one-stage methods~\cite{FAOA,RealTime,ReSC-Large,image_captioning,Zero-shot,onestage1}. Despite the significant achievements, the VG approaches suffer from some limitations, such as failing to capture the detailed feature representation accurately, resulting in a lack of discrimination between fine-grained objects with reference expressions shown in Fig.~\ref{fig:failurecases} (b), and detecting irrelevant or incorrect targets without understanding the whole context shown in Fig.~\ref{fig:failurecases} (c).
\par To address the above issues, this paper proposes a novel iterative robust visual grounding (IR-VG) approach with masked reference based centerpoint supervision. The approach first constructs two new robust VG datasets and proposes a multi-stage false-alarm sensitive decoder (MFSD) module to handle the case when there is no target object from the textual expression, avoiding generating false alarms. Secondly, a new masked reference based centerpoint supervision (MRCS) module is proposed to capture the fine-grained feature and enhance the localization capacity from the given reference expressions. Finally, an iterative multi-level vision-language fusion (IMVF) module is leveraged to fuse multi-level visual and textual information that are crucial for vision-language understanding.
\par The contributions of this paper are summarized as follows: firstly, the proposed approach handles the false-alarm issue in VG task for the \textbf{first time} by constructing two new robust VG benchmarks and introducing a multi-stage false-alarm sensitive decoder (MFSD) module. Secondly, a new masked reference based centerpoint supervision (MRCS) module is proposed to achieve much more accurate fine-grained feature and better localization capacity from fully visual-textual comprehension. Lastly, the iterative multi-level vision-language fusion (IMVF) module is introduced to comprehensively fuse multi-level visual and textual information for better vision-language understanding and alignment. Extensive experiments on five {\it regular} VG benchmarks and two newly constructed {\it robust} VG benchmarks demonstrate the effectiveness of the proposed approach, achieving above \textbf{10\%} improvement on robust datasets.
\section{Related Work}

\noindent
\textbf{Visual Grounding.} The Visual Grounding task is an important problem in computer vision that aims to localize an object within an image based on a given language reference expression. The existing approaches typically extend the object detection framework, such as YOLOV3~\cite{YOLOV3}, Faster-RCNN~\cite{Faster-RCNN}, RetinaNet~\cite{RetinaNet}, CenterNet~\cite{CenterNet}, and DETR~\cite{DETR}, by incorporating a visual-linguistic fusion module. These approaches can be categorized into two main categories: \textit{two-stage methods}~\cite{CMN, NMTree, Ref-NMS,Two-Branch, MAttNet} and \textit{one-stage methods}~\cite{FAOA,RealTime,ReSC-Large,image_captioning,Zero-shot,onestage1}. Two-stage approaches, including CMN~\cite{CMN}, NMTree~\cite{NMTree} and RefNMS~\cite{Ref-NMS}, Two-branch Network~\cite{Two-Branch} and MAttNet~\cite{MAttNet}, utilize an object detector to generate region proposals and then use textual descriptions to select the highest scoring proposal in the second stage. However, this approach can be computationally expensive due to the large number of proposals, and the matching process for each proposal may slow down the inference speed. On the other hand, one-stage approaches~\cite{FAOA,RealTime,ReSC-Large,image_captioning,Zero-shot,onestage1,hu2016segmentation} directly incorporate the linguistic context into visual features to predict the object's location, without generating region proposals. Although one-stage approaches are simple and efficient, they typically rely on pointwise feature representations, which may not be flexible enough to achieve a global context understanding from the vision-language information. Recently, \textit{transformer-based} Visual Grounding approaches have gained popularity due to their attention capacity and efficiency. For instance, TransVG~\cite{TransVG} captures intra- and inter-modal contexts using transformers in a uniform manner, while VLTVG~\cite{vltvg} builds discriminative feature maps and detects the target object through a multi-stage decoder.

\noindent
\textbf{Robustness in Visual Grounding.} Recent studies have explored CNN robustness in various benchmarks~\cite{Robust1}~\cite{Robust2}, and some works have evaluated and improved CNN robustness for practical applications~\cite{Robust3}~\cite{Robust4}~\cite{Robust5}~\cite{RRIS}. RefSegformer~\cite{RRIS} incorporates negative sentence inputs to handle false-alarm issues in referring segmentation tasks. However, to the best of our knowledge, no existing benchmarks or approaches have explored the robustness of the Visual Grounding task. In practice, existing approaches often fail to generate accurate targets when an irrelevant or inaccurate language expression is given. Therefore, this paper takes a further step by proposing a new iterative \textbf{{\it robust}} VG framework and building two robust VG datasets to address this research problem. It is important to note that, within the context of this paper, the term ``\textbf{robust}'' refers to the ability of the proposed method to produce accurate results and avoid false-alarm predictions even when provided with irrelevant and incorrect expressions. 

\noindent
\textbf{Multi-modal Transformer.} Vision transfomer~\cite{DETR,dosovitskiy2020image,li2023transformer,zhang2022eatformer,zhang2023rethinking,li2023tube,zhou2022transvod,li2023panopticpartformer++,li2022videoknet,cheng2021maskformer,xu2022fashionformer,swin} has a a wide range of application, including detecction, representation learning, and segmentation. Recent works~\cite{mdetr,zhu2021uni,zhu2022uni,wu2023open,wu2023betrayed} unify different modal inputs and outputs, mainly representation learning, open vocalbulary, and large language models. For visual grounding, recent works~\cite{RRIS,yang2021lavt,kim2022restr,TransVG} also adopt multu-modal transformer framework, our method belong to this scope. In partilcaur, we pay more attention on the robustness and fine-grained supervision design.

\section{Method}

\begin{figure*}[htbp]
\centering
\includegraphics[scale=0.34]{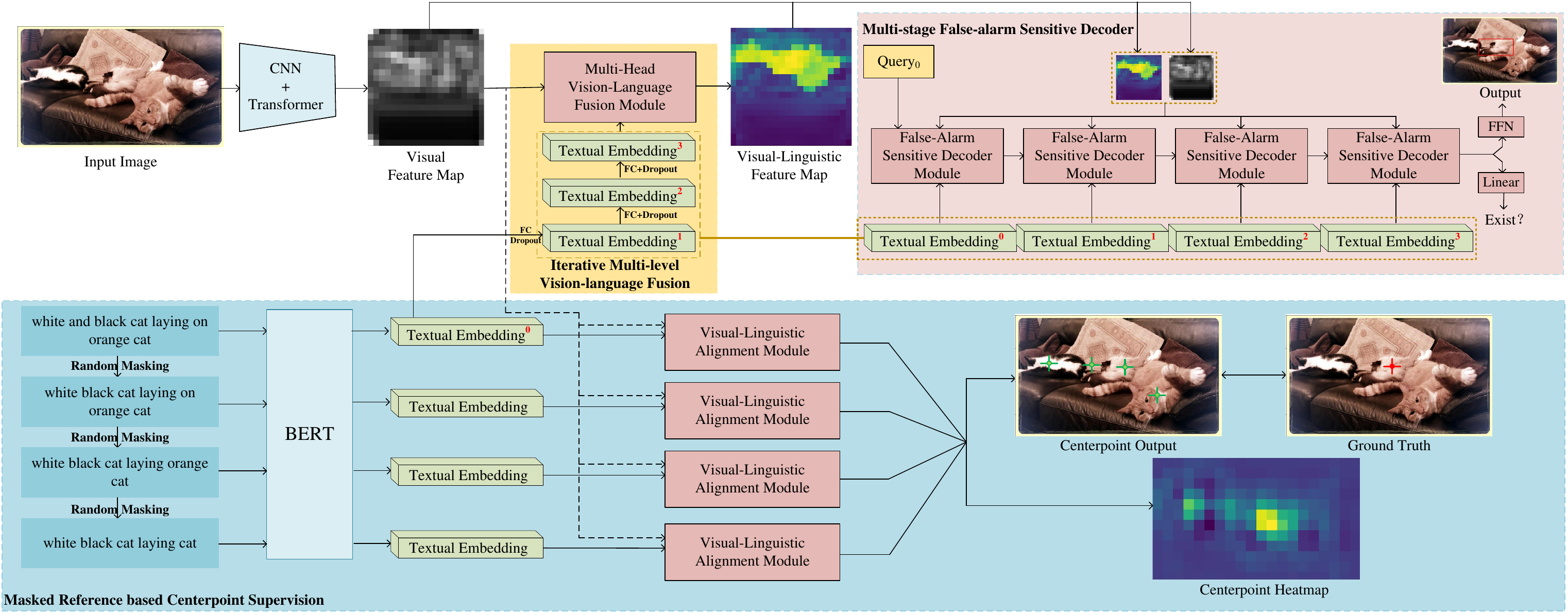}
\caption{An overview of our proposed IR-VG framework, which comprises Masked Reference based Centerpoint Supervision, Iterative Multi-Level Vision-Language Fusion, and Multi-Stage False-Alarm Sensitive Decoder.}
\label{fig:maingraph}
\end{figure*}

In this section, we present the architecture of the proposed robust VG pipeline and its components. Fig.~\ref{fig:maingraph} illustrates the pipeline.
In this section, we present the architecture of the proposed robust VG pipeline and its components. Fig.~\ref{fig:maingraph} illustrates the pipeline, where the image and corresponding language description are processed separately to obtain different feature embeddings in two distinct branches.
% \vspace{-0.2cm}
\subsection{Masked Reference based Centerpoint Supervision}

\par \textbf{Motivation.} Existing VG approaches suffer from inadequate visual-linguistic feature representation, insufficient fine-grained feature representation, and poor localization capacity, leading to the detection of irrelevant or inaccurate objects. To address these issues, we propose the masked reference based centerpoint supervision (MRCS) approach, as illustrated in Fig.~\ref{fig:maingraph}. MRCS comprises three parts: masked reference augmentation, visual-linguistic alignment, and centerpoint supervision. This approach aims to enhance context understanding from the whole image and improve the accuracy of object detection in VG tasks.
\par \noindent \textbf{Masked reference augmentation.} As illustrated in the down-left part of Fig.~\ref{fig:maingraph}, we propose a text augmentation approach to generate diversified textual information given an input language expression. We employ the NLTK~\cite{nltk} tokenization strategy to extract lexical properties for each word, followed by masking one word in the text according to the well-designed rules (shown in the supplementary materials). This masking process is repeated at most three times, achieving one full text and three masked texts in total. BERT~\cite{Bert} is then utilized to generate different textual embeddings for these sentences.
\par It is important to note that we prioritize the masking of lexical words differently based on their semantic significance. Prepositions, conjunctions, and qualifiers are masked first, as they generally have minimal impact on the sentence's meaning. If these types of words are absent, the module masks auxiliaries, pronouns, and numbers, which can partially affect the sentence's semantics. Finally, the module masks adjectives and verbs, which are critical for the sentence's meaning. If there is only one non-noun word or only nouns remaining in the sentence, no further masking is performed. More specific rules will be shown in the supplementary materials.
\par
\begin{figure}[tbp]
  \centering
  \includegraphics[scale=0.33]{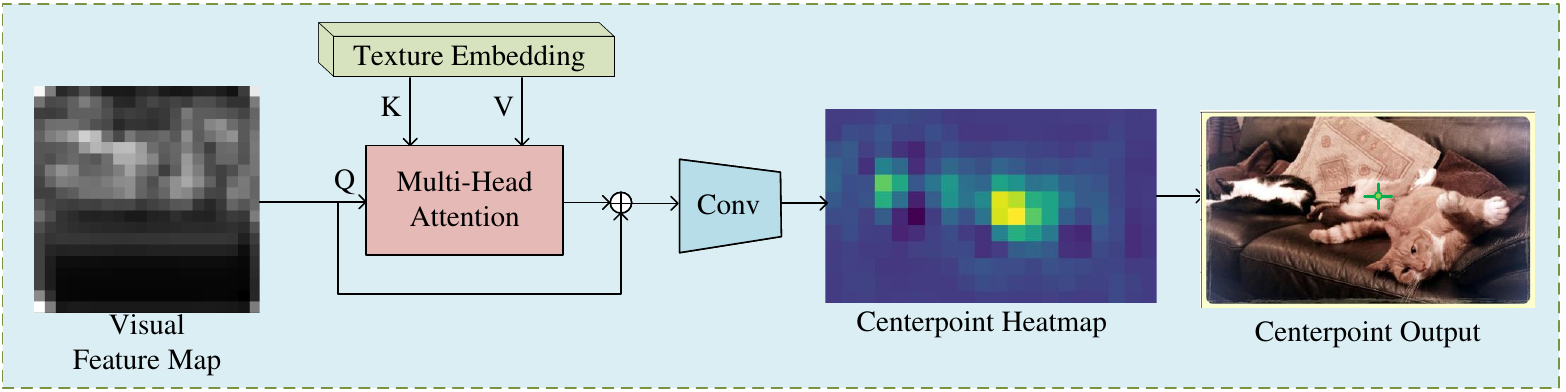}
  \caption{The architecture of visual-linguistic alignment module.}
  \label{fig:KP_Reg}
\end{figure}

 \noindent \textbf{Visual-linguistic alignment.} The proposed model, illustrated in Fig.~\ref{fig:KP_Reg}, incorporates a visual-linguistic alignment module with two consecutive MHA layers. The visual feature map $F_{v}$ is input as the {\it Query}, and the textual embeddings are input as {\it Key} and {\it Value} to the first MHA. This process produces an enhanced feature map that gathers relevant semantic information from the corresponding linguistic representation. Subsequently, the enhanced feature map undergoes another MHA operation that performs self-attention on the visual features to encode the involved visual contexts. The features from the two MHAs are element-wisely summed in a residual manner for the centerpoint supervision component.
The goal of these two MHA operations is to encode the related descriptions into the visual feature and enhance the visual context information from the whole image. The features from the two MHAs are element-wisely summed in a residual manner for the centerpoint supervision component.
As shown in Fig.~\ref{fig:KP_Reg}, the textual embeddings from the language branch and the visual feature map from the image branch will be input to the visual-linguistic alignment module based on two consecutive multi-head attention (MHA) layers. Specifically, we input the visual feature map $F_{v}$ as the {\it Query}, and textual embeddings as {\it Key} and {\it Value} into the first MHA layer, where enhanced feature map will be achieved by collecting the relevant semantic information from the corresponding linguistic representation. The enhanced feature will then again be processed through another MHA operator that performs self-attention for the visual features to encode the involved visual contexts. The two consecutive MHA operations try to encode the related descriptions into the visual feature and enhance the visual context information from the whole image. The features from the two MHAs will be element-wisely summed together in a residual manner, which will be employed in the keypoint supervision part.
\begin{figure}[tbp]
  \centering
  \includegraphics[scale=0.36]{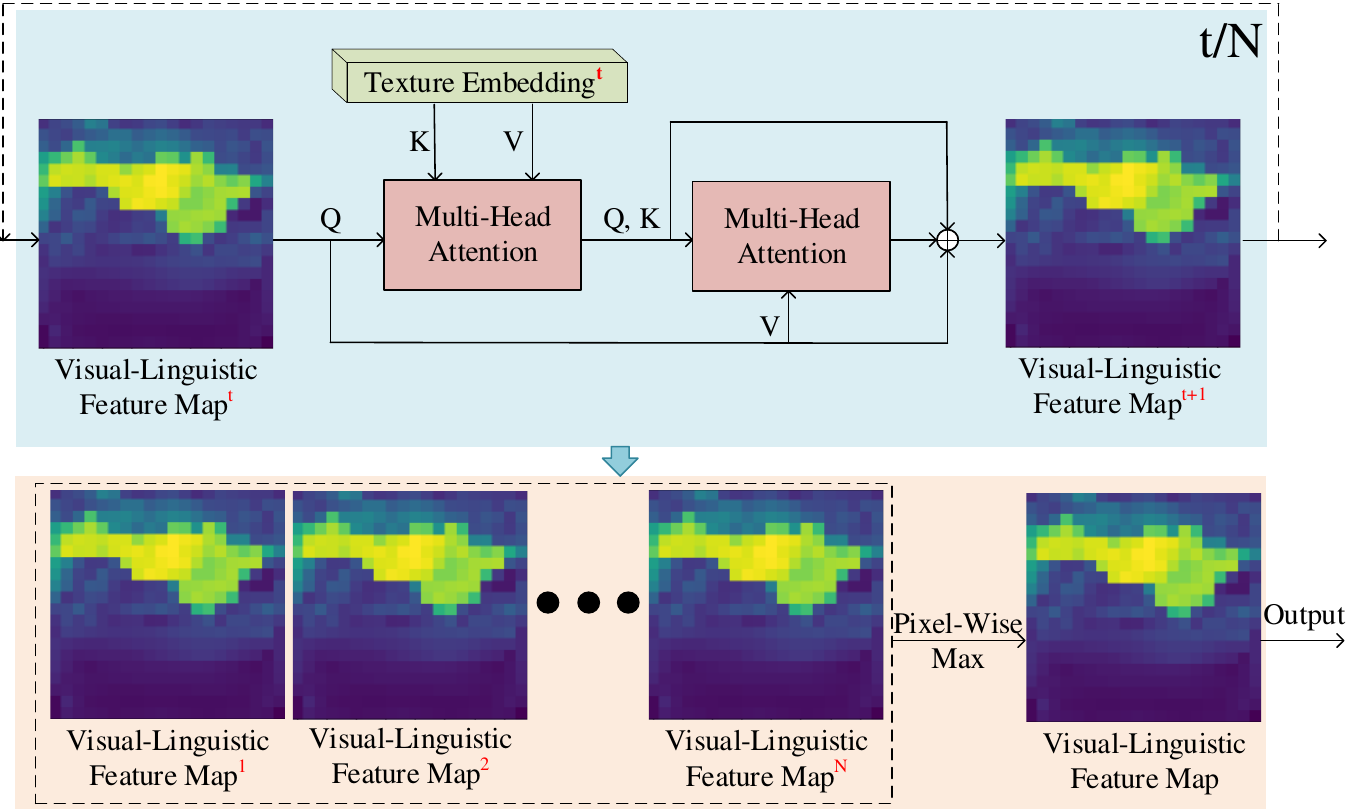}
  \caption{The architecture of IMVF.}
  \label{fig:V-L_Fusion}
\end{figure}
\begin{figure}[tbp]
  \centering
  \includegraphics[scale=0.40]{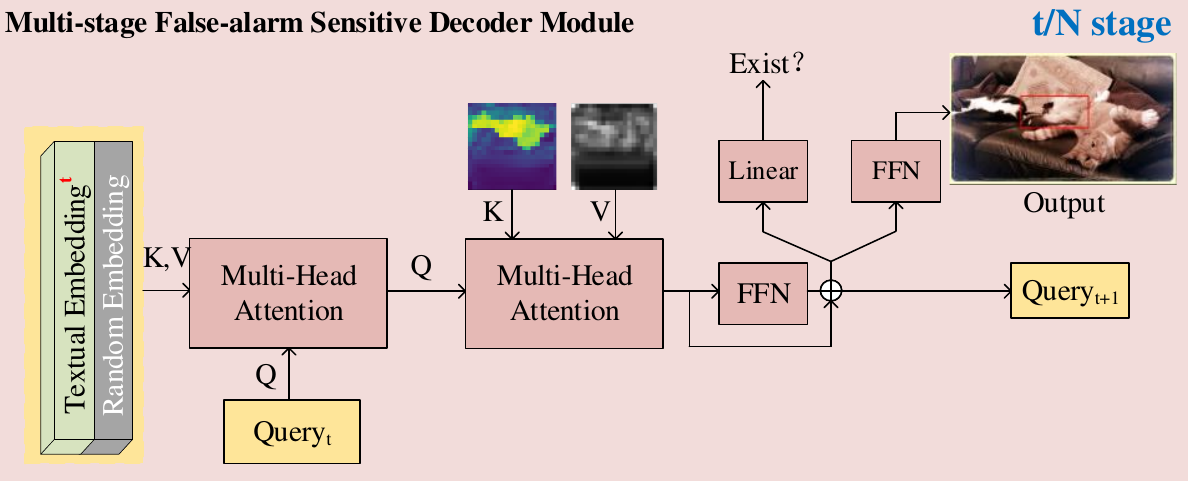}
  \caption{The architecture of MFSD.} 
  \label{fig:falsealarm}
\end{figure}

\noindent \textbf{Centerpoint supervision.} To obtain the final centerpoint heatmaps, the summed feature map obtained from each language expression is processed through two consecutive convolutional layers. Multiple centerpoint heatmaps (one from the full text and three from the masked text) are then fused by performing a {\it maxpooling} process, with the centerpoint coordinates determined by performing a {\it argmax} operation on the resulting heatmap. The cross entropy loss is then utilized as the supervision loss between the centerpoint heatmap and the corresponding ground truth, given by $\mathcal{L}_\text{key} = \text{CELoss}(y, \hat{y}_{i})$, where $\text{CELoss}(\cdot, \cdot)$ is the cross entropy loss, $\hat{y}_{i}$ is the predicted centerpoints, and $y$ is the centerpoint ground truth that is obtained from the center point of each ground truth box. 

\subsection{Iterative Multi-level Vision-language Fusion} 

\noindent
\textbf{Motivation.} Through empirical analysis, we have observed that the visual-textual misunderstanding issue arises due to inadequate and poor textual embeddings in the multi-head vision-language fusion module, which incorporates different visual features from various stages. To address this challenge, we propose a multi-level textual feature enhancement (MTFE) module that enhances textual embeddings from low-level to high-level, analogous to the image feature extraction branch. The module extracts multi-level textual information from the entire sentence, resulting in more comprehensive and robust textual embeddings. 
\par \noindent \textbf{Multi-level textual feature enhancement.} The MTFE module improves textual embedding representation by performing two consecutive fully-connected layers with 768 nodes in each stage. Specifically, as highlighted with yellow color in Fig.~\ref{fig:maingraph}, the IMVF comprises four stages, and each stage contains an MTFE module. The MTFE module consists of two fully connected layers and a corresponding dropout layer with a 0.1 ratio, aimed at obtaining multi-level textual features that match the multi-level visual features. This enables the model to focus on different key descriptions in the referring expressions and obtain more complete and reliable features for the referred object.
\par \noindent \textbf{Iterative multi-level vision-language fusion.} Fig.~\ref{fig:V-L_Fusion} illustrates the IMVF module, which is based on MHA and consists of four iterative stages. Each stage includes two MHA layers. The first layer uses the visual feature map $F_{v} \in \mathcal{R}^{C\times H \times W}$ as the {\it Query} and the textual embeddings $F_{l} \in \mathcal{R}^{C \times L}$ from the multi-level textual feature enhancement module as the {\it Key} and {\it Value}. Multi-head cross-attention enables the comprehensive incorporation of textual information into the visual feature map $F_{g} \in \mathcal{R}^{C \times H \times W}$. In the second layer, $F_{g}$ serves as both the {\it Query} and {\it Key}, while $F_{v}$ serves as the {\it Value}. This self-attention operator allows the model to gather crucial context features for the referred object based on the textual descriptions provided, and the final feature is $F_{c} \in \mathcal{R}^{C \times H \times W}$. We sum the $F_{v}$, $F_{g}$, and $F_{c}$ element-wisely to obtain the final visual feature map $F_{m}$. In each iteration, the $i$-th visual feature map $F^{i}_{m}$ becomes the initial feature map (i.e., $F^{i+1}_{v}$). Our experiments include four iterations, and we use element-wise {\it max} strategy to obtain the final fusion feature $F=\text{max}(F^{1}_{m}, F^{2}_{m}, F^{3}_{m}, F^{4}_{m})$. Actually, other fusion strategies can also be considered. We experimentally find that element-wise summation or product achieves inferior performance than the proposed strategy.

\subsection{Multi-stage False-alarm Sensitive Decoder}

\noindent
\textbf{Motivation.} The current SOTA approaches in VG task assume that the language expressions are precisely matched with the visual image. However, this assumption may not hold in practical applications. Specifically, when an inaccurate or irrelevant text expression is provided, the existing SOTA VG approaches~\cite{TransVG,vltvg} often generate false-alarm results. To address this issue, we introduce several \textbf{robust} VG datasets (described in Sec. 4) and propose a new multi-stage false-alarm sensitive decoder (MFSD) module.
\par \noindent \textbf{Multi-stage false-alarm sensitive decoder.} As shown in Fig.~\ref{fig:falsealarm}, the MFSD module consists of several iterative stages, each contains two consecutive multi-head attention (MHA)~\cite{MHA} layers. In the first stage, we randomly initialize a series of learnable queries. To handle the false-alarm case, we introduce a random embedding with the same size as textual embedding from the IMVF module. We concatenate the textual embedding and the random embedding in the batch dimension, termed as {\it mixture embedding}. For the first MHA layer, the learnable queries serves as {\it Query}, and the mixture embedding acts as {\it Key} and {\it Value}. With this layer, the textual embedding can be more easily attended to the target tokens, thus achieving enhanced textual embedding. For the second MHA layer, the enhanced textual embedding is treated as {\it Query}, and the visual-linguistic feature map from the IMVF module as well as the visual feature map $F_v$ are employed as {\it Key} and {\it Value}. Through the second MHA layer, the textual information can be comprehensively fused with the visual feature map to achieve an enhanced vision-language feature, which is then taken into a feed-forward network (FFN). We fuse the enhanced vision-language feature and the feature from the second MHA in a residual manner, termed as R\_feature, which serves as the {\it Query} in the next iteration. Then, R\_feature is taken into two decoupled heads: one for classification to indicate whether there exists false-alarm result, and another for regression to generate the predicted bounding boxes (bbox). 
Specifically the classification loss $\mathcal{L}_{\text{cls}}$ and the regression loss $\mathcal{L}_{\text{reg}}$ are defined as,
\begin{equation}
    \mathcal{L}_{\text{cls}} = \sum_{t=1}^{N}\sum_{i=1}^{K}\text{CELoss}(y^{t}, \hat{y}^{t}_{i}),
\end{equation}
\begin{equation}
     \mathcal{L}_{\text{reg}} = \sum_{t=1}^{N}\sum_{i=1}^{K}\lambda_\text{GIOU}\mathcal{L}_{\text{GIOU}}(b^{t},\hat{b}^{t}_{i}) + \lambda_{\text{L1}}\mathcal{L}_{\text{L1}}(b^{t},\hat{b}^{t}_{i}),
\end{equation}
where $\text{CELoss}(\cdot, \cdot)$, $\mathcal{L}_{\text{GIOU}}(\cdot, \cdot)$ and $\mathcal{L}_{\text{L1}}(\cdot, \cdot)$ are the cross entropy loss, GIOU loss~\cite{GIOU} and L1 loss, respectively. $y^{t}$ and $\hat{y}^{t}_{i}$ denote the ground truth label and predicted result in $t$-th iteration. Similarly, $b^{t}$ and $\hat{b}^{t}_{i}$ denote the ground truth bbox and predicted bbox. $t$ denotes the $t$-th iteration, and $i$ represents the $i$-th bbox. $\lambda_{\text{GIOU}}$ and $\lambda_{\text{L1}}$ are empirically adjusted, here we set them as 3 and 7 by default for all the following experiments.

\subsection{Details of Determining False-alarm Detection of the Previous Methods}

In the previous methods, we follow the same rules as ours to obtain the false alarm. Firstly, we achieve the top1 scoring box as the final prediction box. Then, we calculate the IOU value with the ground truth box. If the IOU value is greater than 0.5, we consider it a true positive, otherwise, we treat it as a false positive. However, the proposed method differs in that it combines the top1 scoring box and its existing result (exist or non-exist) to achieve the final prediction box. During our experiments, we attempted to add an irrelevant text reference head to some previous networks, such as VLTVG~\cite{vltvg} but the results were inferior to their baselines. It may not be fair to compare these results in the paper, thus we do not show these results.

\subsection{Training Loss}

\par In the training stage, the proposed VG framework is trained end-to-end using the aforementioned losses. The overall loss function for the proposed framework is $\mathcal{L} = \mathcal{L}_{\text{cls}} + \lambda_{\text{reg}}\mathcal{L}_{\text{reg}} + \lambda_{\text{key}}\mathcal{L}_{\text{key}}$ as follows, where $\mathcal{L}_{\text{cls}}$, $\mathcal{L}_{\text{reg}}$, and $\mathcal{L}_{\text{key}}$ denote the classification loss, regression loss and centerpoint loss, respectively. $\lambda_{\text{reg}}$ and $\lambda_{\text{key}}$ are introduced to balance the above losses.We empirically set $\lambda_{\text{reg}}$ and $\lambda_{\text{key}}$ as 2 and 5 by default. 
\begin{equation}
    \mathcal{L} = \mathcal{L}_{\text{cls}} + \lambda_{\text{reg}}\mathcal{L}_{\text{reg}} + \lambda_{\text{key}}\mathcal{L}_{\text{key}},
\end{equation}

\section{Experiment}

\subsection{Experimental Settings}

\par \noindent \textbf{Datasets.} To comprehensively verify the effectiveness of the proposed robust VG approach, we evaluate it on two types of datasets: the regular VG datasets and the robust VG datasets.
\par \noindent \textbf{Regular VG datasets.} We evaluate our proposed approach on five regular VG datasets, including the RefCOCO~\cite{RefCOCO}, RefCOCO+~\cite{RefCOCO}, RefCOCOg~\cite{RefCOCOg}, ReferItGame~\cite{ReferItGame}, and Flickr30k~\cite{Flickr30k}. The RefCOCO datasets series, including RefCOCO, RefCOCO+, and RefCOCOg, are three commonly used benchmarks for visual grounding, the images used in these datasets are collected from the train2014 set of MSCOCO dataset. Specifically, the RefCOCO dataset contains 19,994 images, 50,000 reference objects, and a total of 142,210 reference expressions. Among them, 120,624 reference expressions are used as the training set, 10,834 as the validation set, 5657 and 5095 expressions for test A and test B, respectively. The RefCOCO+ dataset provides 19,992 images with 49,856 reference objects and 141,564 reference expressions. Similar to RefCOCO, RefCOCO+ is also divided into training, validation, test A, and test B sets, with 120,191, 10,758, 5,726, and 4,889 reference expressions in these datasets. RefCOCOg contains a total of 25,799 images, 49,822 objects, and 95,010 reference expressions. Compared to the first two datasets, most of the expressions in RefCOCOg have longer sentences and more complex statement structures. RefCOCOg contains two sub-datasets, RefCOCOg-google and RefCOCOg-umd. Since the former dataset does not provide a test set, we mainly use the RefCOCOg-umd dataset. ReferItGame contains 20,000 images, which are collected from the SAIAPR-12 dataset. This dataset has a total of 120,072 reference expressions and is divided into a training set with 54,127 reference expressions, a validation set with 5,842 reference expressions, and a test set with 60,103 reference expressions. Flickr30k contains 31,783 images and 427,000 reference expressions. We divide the training, validation, and test sets using the same ratio as the previous work.
% RefCOCO~\cite{RefCOCO}, RefCOCO+~\cite{RefCOCO}, RefCOCOg~\cite{RefCOCOg}, ReferItGame~\cite{ReferItGame}, Flickr30k~\cite{Flickr30k}, under the experimental settings identical to those of \cite{vltvg}
\par \noindent \textbf{Robust VG Datasets} We construct two robust VG datasets based on the existing benchmarks RefCOCOg and ReferItGame, termed RefCOCOg\_F and ReferItGame\_F. The train set of our robust VG datasets contains two parts of data, the first part is the train set of the original dataset, while the second part is a random matching dataset, which destroys the correspondence between the image information and the language descriptions. 
Specifically, for each target on the image, we select one description that is different from its original one among all the text descriptions in the dataset, thus building a dataset where the image is with irrelevant or inaccurate descriptions. During training, the ratio of these two parts of data is 1:1. The test set of our robust VG datasets also consists of two parts of data, the first part is the test set of the original dataset while the second part is the manually modified robust VG dataset, which requires manual intervention to modify some keywords in the descriptions, thus modifying the semantics of the descriptions and building a more difficult dataset. For instance, we manually modify the expression “The man in white T-shirt is riding a bike” to “The man in blue T-shirt is riding a bike”. Specifically, the test set of the RefCOCOg\_F dataset contains 2000 pairs of false-alarm data and 9602 pairs of regular data that are from the original RefCOCOg test set. The test set of the ReferItGame\_F dataset contains 1000 pairs of false-alarm data and 9000 pairs of regular data that are randomly sampled from the test set of the original ReferItGame dataset. 
\par Specifically, the data combination method of the random matching dataset is to randomly replace the description in each group of data in the training set with a random other description in the dataset to construct false-alarm data. Of course, the description of the same image will not be selected to avoid the existence of the target corresponding to the ran71 dom description on the image. It can be observed that the probability of the existence of the target corresponding to the description on the image is very low for the false alarm data formed by this random selection description method. 
\par
\begin{figure}[tbp]
\centering
\includegraphics[scale=0.40]{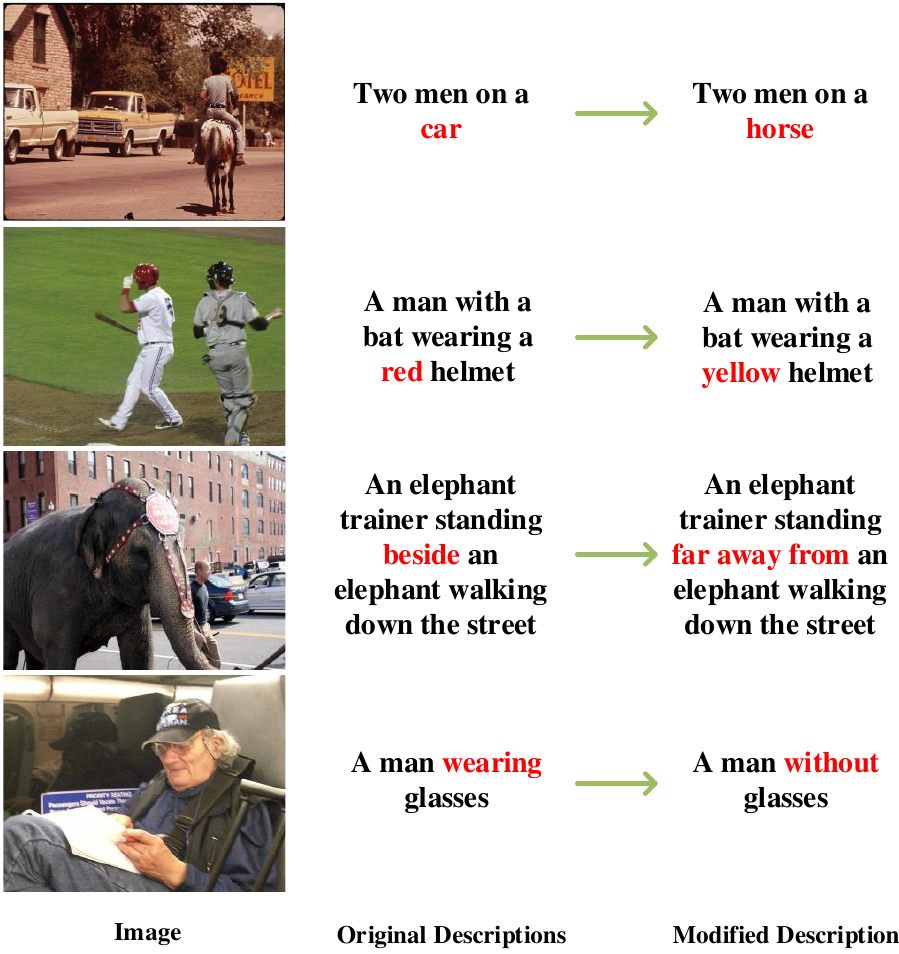}
\caption{Example of manually modified false-alarm data.}
\label{fig:FA_Modified}
\end{figure}

 We build the manually modified robust VG dataset by manually modifying some keywords in the description. In general, we mainly modify words from the following perspectives. First, modifying key nouns can greatly change the semantics of words, thus generating false alarm data. For example, modify ”Two men on a horse” to ”Two men on a car” (as shown in the first row of Fig.~\ref{fig:FA_Modified}). Second, modifying key adjectives can also change the description semantics. For example, modify ”A man with a bat wearing a red helmet” to ”A man with a bat wearing a yellow helmet” (as shown in the second row of Fig.~\ref{fig:FA_Modified}). Third, modify words in the text that relate to spatial location can mismatch the original target with the newly generated text. For example, modify ”An elephant trainer standing beside an elephant walking down the street” to ”An elephant trainer standing far away from an elephant walking down the street” (as shown in the third row of Fig.~\ref{fig:FA_Modified}). Fourth, changing the words corresponding to some fine-grained features can generate false-alarm data. For example, modify ”A man wearing glasses” to ”A man without glasses” (as shown in the fourth row of Fig.~\ref{fig:FA_Modified}). Experiments show that our pro95 posed IR-VG is effective for all four types of false alarm data.
\par \noindent \textbf{Implementation Details.} Consistent with SOTA approaches such as TransVG~\cite{TransVG} and vltvg~\cite{vltvg}, our proposed method employs ResNet101~\cite{ResNet} as the backbone, augmented with 6 transformer layers in the image feature extraction branch, initialized using weights from DETR~\cite{DETR}. The textual embedding extraction branch is initialized with BERT~\cite{Bert}, while the parameters of other components use Xavier scheme~\cite{Xavier} initialization. We resize all images to $640\times640$ and fill them with black to form a square. We perform experiments using PyTorch, a 3090ti GPU, a batch size of 16, and run training for 90 epochs using the AdamW optimizer with a learning rate of $3\times10^{-4}$ and a weight decay of $1\times10^{-4}$.
\begin{figure}[tbp]
\centering
\includegraphics[scale=0.40]{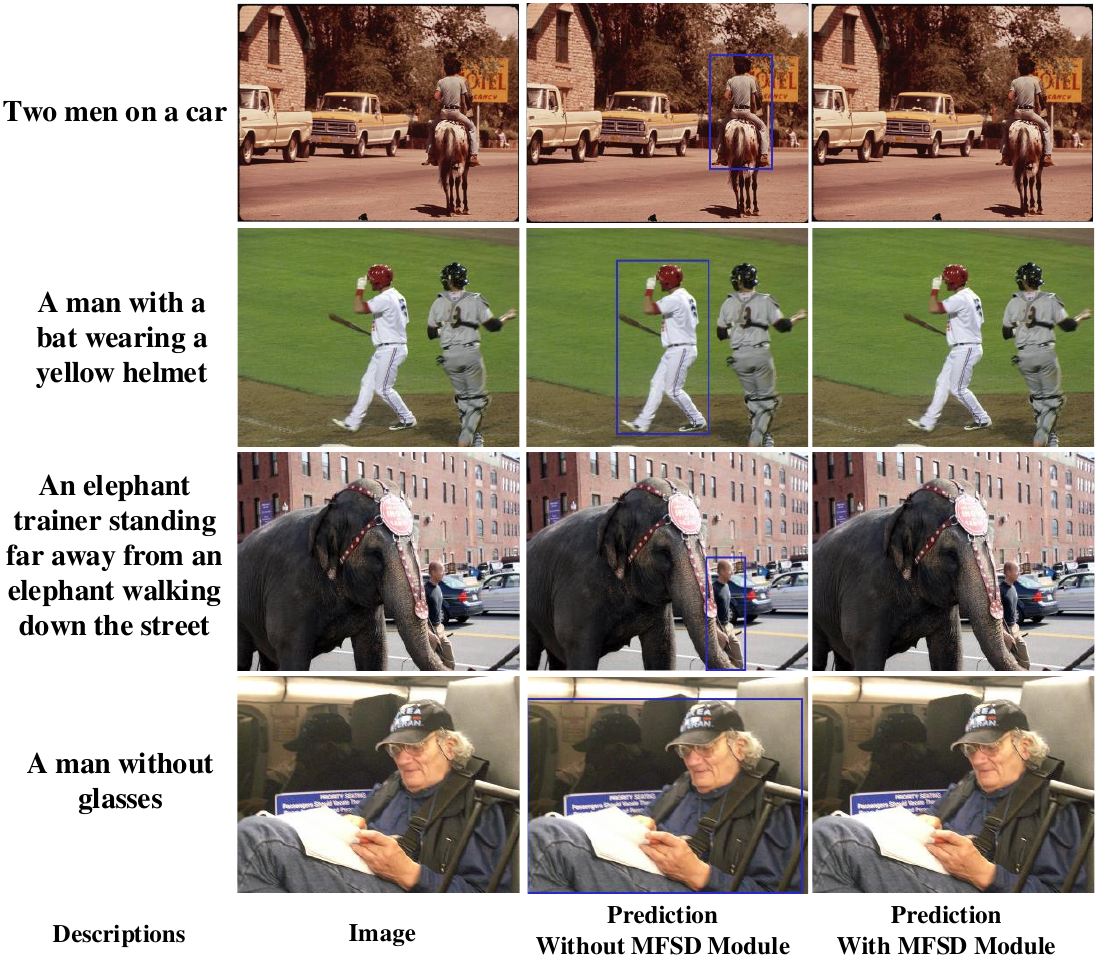}
\caption{ Visualization of the MFSD module.}
\label{fig:FA_vis}
\end{figure} 
\begin{table*}[tbh]
\begin{center}
\setlength{\tabcolsep}{2.0mm}{
\begin{tabular}{|c|c c c|c c c|c c|c|c|}
\hline
 \multirow{2}{*}{Method} &
      \multicolumn{3}{c|}{RefCOCO}  & \multicolumn{3}{c|}{RefCOCO+} & \multicolumn{2}{c|}{RefCOCOg} &\multicolumn{1}{c|}{ReferItGame}  &\multicolumn{1}{c|}{Flickr30k} \\
                                   &$\text{val}$   &$\text{testA}$  &$\text{testB}$  
    &$\text{val}$   &$\text{testA}$  &$\text{testB}$ 
    &$\text{val-u}$   &$\text{test-u}$
    &$\text{test}$
    &$\text{test}$  \\
\hline
  CMN~\cite{CMN} &-  &71.03 &65.77 &-  &54.32 &47.76 &-  &- &28.33  &-\\
  VC~\cite{VC} &-  &73.33 &67.44 &-  &58.40 &53.18 &-  &- &31.13  &-\\
  NMTree~\cite{NMTree} &76.41  &81.21 &70.09 &66.46  &72.02 &57.52 &65.87  &66.44 &-  &-\\
  Ref-NMS~\cite{Ref-NMS} &80.70  &84.00 &76.04 &68.25  &73.68 &59.42 &70.55  &70.62 &-  &-\\
  FAOA~\cite{FAOA} &72.54 &74.35 &68.50  &56.81 &60.23 &49.60  &61.33 &60.36  &60.67 &68.71\\
  LBYLNet~\cite{LBYL-Net} &79.67  &82.91 &74.15 &68.64  &73.38 &59.49 &-  &- &67.47  &-\\
  TransVG~\cite{TransVG} &81.02  &82.72 &78.35 &64.82  &70.70 &56.94 &68.67  &67.73 &70.73  &79.10\\
  VLTVG~\cite{vltvg} &84.77  &87.24 &80.49 &74.19  &78.93 &65.17 &76.04  &74.98 &71.98  &79.84\\
  IR-VG (Ours) &\textbf{86.82}  &\textbf{88.75} &\textbf{82.60} &\textbf{76.22}  &\textbf{80.75} &\textbf{67.33} &\textbf{77.86}  &\textbf{76.24} &\textbf{74.03}  &\textbf{81.45}\\
\hline
\end{tabular}}
\end{center}
\caption{Comparisons with SOTA visual grounding methods.}
\label{result}
\end{table*}
% \vspace{-0.1cm}
\begin{table}
\begin{tabular}{|c|c c|c c|}
\hline
  \multirow{2}{*}{Methods} & \multicolumn{2}{c|}{RefCOCOg\_F~~}  & \multicolumn{2}{c|}{ ReferItGame\_F}
  \\
  &$R_\text{fad}$ &$R_\text{mix}$
  &$R_\text{fad}$ &$R_\text{mix}$
  \\
  \hline
  CMN~\cite{CMN} &~27.10&~65.10 &~24.75 &~21.41~  \\
  VC~\cite{VC} &~42.45 &~68.85 &~31.03 &~25.69~  \\
  SSG~\cite{SSG} &~34.15  &~61.25 &~32.44 &~46.43~ \\
  Ref-NMS~\cite{Ref-NMS} &~43.90  &~62.40 &~41.39 &~48.15~\\
  ReSC-Large~\cite{ReSC-Large} &~37.35  &~60.55 &~32.54 &~59.89~ \\
  LBYLNet~\cite{LBYL-Net} &~45.40  &~63.32 &~45.40 &~60.57~ \\
  IR-VG (Ours) &\textbf{~67.32}  &\textbf{~73.61} &\textbf{~69.44} &\textbf{~72.03~}\\
\hline
\end{tabular}
\caption{Comparisons with SOTA approaches on {\it \textbf{robust}} VG datasets.}
\label{false_alarm}
\end{table}
% \vspace{-0.2cm}
\begin{table}
\setlength{\tabcolsep}{0.5mm}{
\begin{tabular}{|c c|c c c|c c c|c c|}
\hline
  \multicolumn{2}{|c|}{Methods}
  &\multicolumn{3}{c|}{RefCOCO}
  &\multicolumn{3}{c|}{RefCOCO+}
  &\multicolumn{2}{c|}{RefCOCOg} 
  \\
  $\text{I}$     &$\text{M}$
  &$\text{val}$   &$\text{testA}$  &$\text{testB}$  
  &$\text{val}$   &$\text{testA}$  &$\text{testB}$ 
  &$\text{val-u}$   &$\text{test-u}$
  \\
  \hline
  - &- &84.77  &87.24 &80.49 &74.19  &78.93 &65.17 &76.04  &74.98\\
  \checkmark &- &85.92  &88.41 &81.77 &75.27  &80.06 &66.33 &77.10  &76.06 \\
  - &\checkmark &85.53  &88.09 &81.23 &75.34  &79.97 &66.18 &77.21  &75.75 \\
  \checkmark &\checkmark &\textbf{86.82}  &\textbf{88.75} &\textbf{82.60} &\textbf{76.22}  &\textbf{80.75} &\textbf{67.33} &\textbf{77.86}  &\textbf{76.24}\\
\hline
\end{tabular}}
\caption{Ablation studies on three benchmarks, "I" and "M" denote IMVF and the MRCS.}
\label{result2}
\end{table}
% \vspace{-0.2cm}
\par \noindent \textbf{Evaluation Metrics.}
For the {\it \textbf{regular}} VG datasdet, following previous works~\cite{TransVG}~\cite{Metric1}, we adopt the commonly used top1 accuracy (acc-1) as the evaluation metric. For the {\it \textbf{robust}} VG dataset, we propose two novel evaluation metrics, i.e., false alarm discovery rate $R_\text{fad}$ with only false-alarm data, and correct rate among the mixed data $R_\text{mix}$ with both false-alarm and regular data, which are defined as, 
%\begin{multicols}{2}
\begin{equation}
    R_\text{fad} = \frac{\text{FA}^\text{acc}}{\text{FA}^\text{all}},~~~
    R_\text{mix} = \frac{\text{FA}^{\text{acc}} + \text{Regular}^\text{acc}}{\text{FA}^\text{all} + \text{Regular}^\text{all}},
\end{equation}
where FA denotes the false-alarm data with irrelevant or inaccurate descriptions, and Regular means the regular data with accurate descriptions. The superscript \textbf{acc} and \textbf{all} represent the number of accurate predictions and the total number of the data. The detailed dataset descriptions, training loss and other experiment implementation details will be shown in the supplementary materials.

\subsection{Comparisons with Existing SOTA Methods}

\par As presented in Tab.~\ref{result}, we evaluate the proposed approach against other SOTA VG methods. Numerically, we improve over the best SOTA approaches by about 2\% in all five benchmarks, indicating the effectiveness of our proposed method.
\par Tab.~\ref{false_alarm} demonstrates the numerical comparisons on the {\it \textbf{robust}} VG datasets. Obviously, we improve over the SOTA approaches by a nontrivial margin in competitive benchmarks of RefCOCOg\_F and ReferItGame\_F. Specifically, on ReferItGame\_F dataset, we achieve about 25\% and 10\% improvement in $R_\text{fad}$ and $R_\text{mix}$ metrics, respectively. It is worth noting that TransVG~\cite{TransVG} and VLTVG~\cite{vltvg} are not included in the comparison because they only provide one predicted bounding box without any extra information to determine whether the target object is a false alarm. As a result, they will definitely generate false-alarm objects when given inaccurate or irrelevant language expressions, which is not a fair comparison. 
\begin{figure}[tbp]
\includegraphics[scale=0.345]{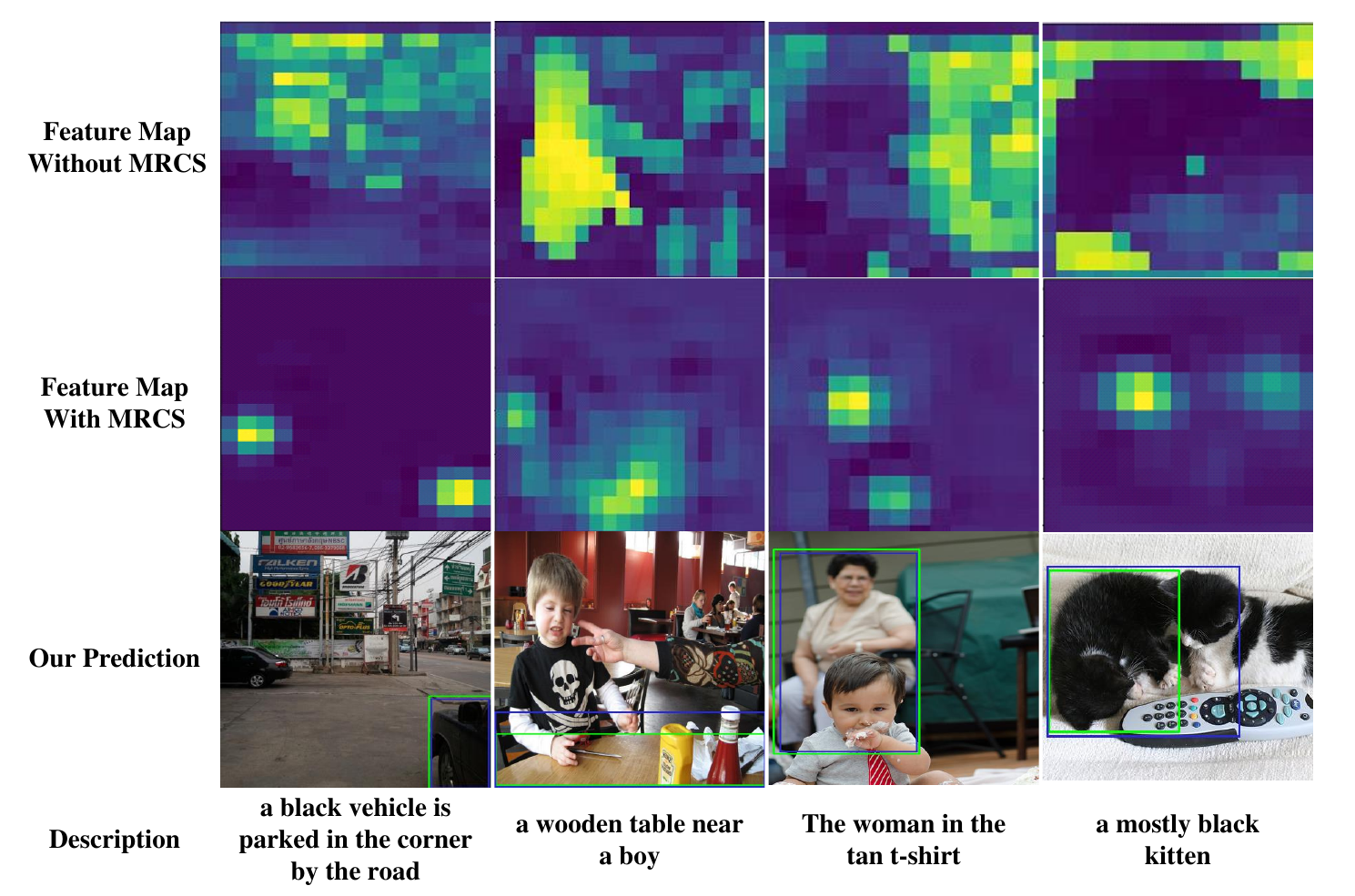}
\caption{Visualization of the visual-linguistic feature map (shown in Fig.~\ref{fig:maingraph}) with/without MRCS module.}
\label{fig:kp_vis}
\end{figure}
\begin{figure}[tbp]
\includegraphics[scale=0.365]{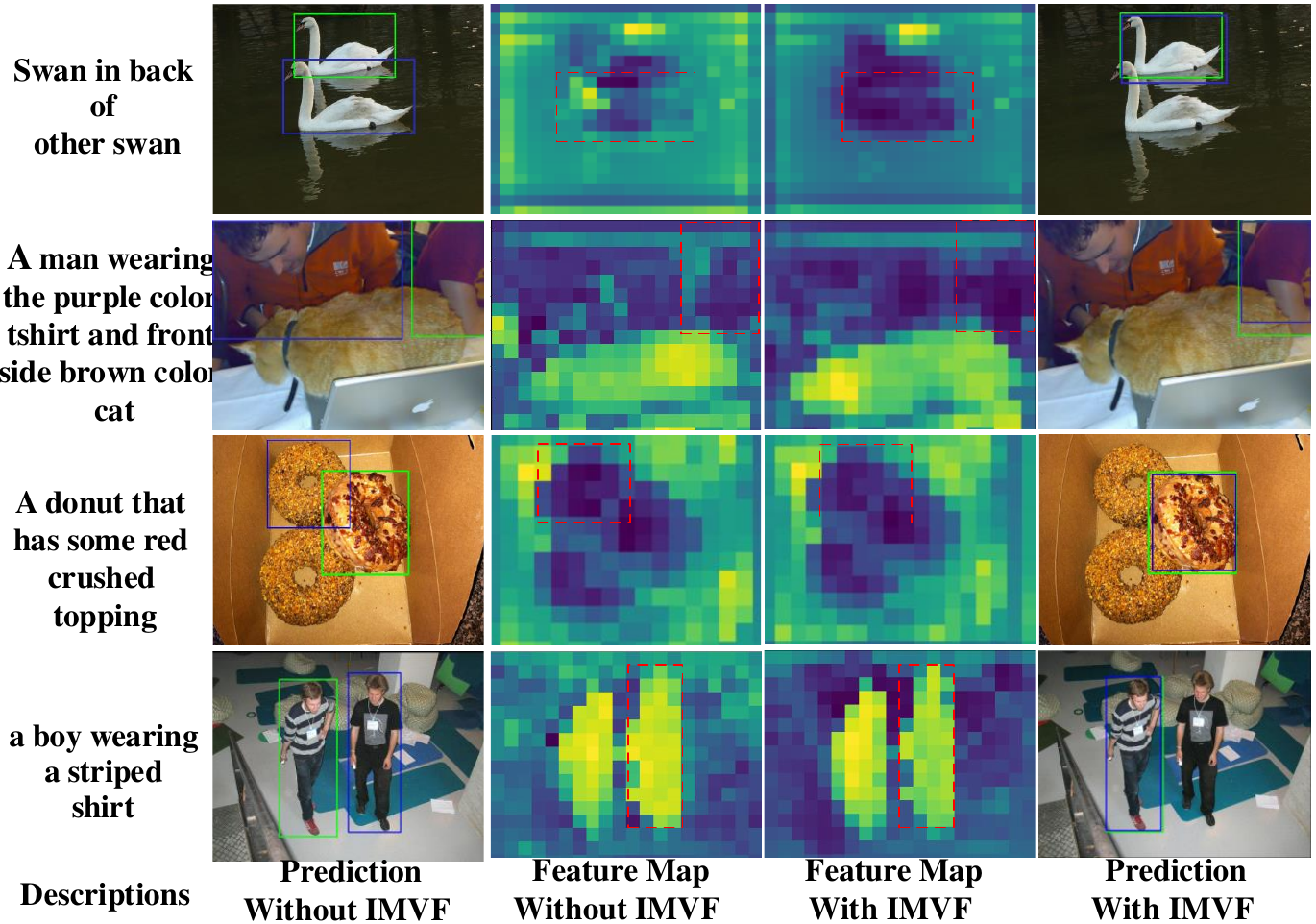}
\caption{Visualization of the visual-linguistic feature map (shown in Fig.~\ref{fig:maingraph}) with/without the IMVF module, especially for the red rectangle areas.}
\label{fig:TA}
\end{figure}
% \vspace{-0.2cm}
\subsection{Ablation Study}

\par \textbf{Numerical Component Analysis.} Tab.~\ref{result2} shows the effectiveness of each component on the {\it regular} VG datasets. The proposed approach outperforms the baseline by 2.1\% top1 accuracy in RefCOCO testB dataset. Specifically, IMVF improves by 1.3\% and MRCS improves by 0.7\%. Similar conclusions can be drawn from other {\it regular} VG datasets. Tab.~\ref{false_alarm} illustrates the effectiveness and robustness of the proposed MFSD module, which achieves a significant improvement on two competitive robust benchmarks. For instance, the MFSD module improves by 25\% and 10\% compared with existing SOTA approaches in $R_\text{fad}$ and $R_\text{mix}$ metrics, respectively.
\begin{table}\scriptsize
  \small
  \centering
  \setlength{\tabcolsep}{0.3mm}{
  \begin{tabular}{|c|c c c|c c c|c c|}
  \hline
  \multicolumn{1}{|c|}{Methods}
  &\multicolumn{3}{c|}{RefCOCO}
  &\multicolumn{3}{c|}{RefCOCO+}
  &\multicolumn{2}{c|}{RefCOCOg} 
  \\
                &$\text{val}$   &$\text{testA}$  &$\text{testB}$  
                &$\text{val}$   &$\text{testA}$  &$\text{testB}$ 
                &$\text{val-u}$   &$\text{test-u}$\\
                \hline
   Baseline~ &~85.92  &~88.41 &~81.77~ &~75.27  &~80.06 &~66.33~ &~77.10  &76.06 \\
   Wo masked~ &~86.57  &~88.52 &~82.26~ &~75.84  &~80.41 &~67.02~ &~77.57  &75.93 \\
  Ours~ &~\textbf{86.82}  &~\textbf{88.75} &~\textbf{82.60}~ &~\textbf{76.22}  &~\textbf{80.75} &~\textbf{67.33}~ &~\textbf{77.86}  &~\textbf{76.24}\\
  \hline
  \end{tabular}}
 \caption{Ablation study on multiple masked strategies. ``Baseline'' denotes the experiment with one full text without centerpoint supervision, ``Wo masked'' denotes the result with one full text and centerpoint supervision, and ``Ours" represents the experiment with MRCS.}
 \label{result_s}
\end{table}
% \vspace{-0.2cm}
\subsection{Rules for masking words in MRCS module.}

When we mask lexical words, we prioritize them differently. We first mask prepositions, conjunctions, and qualifiers because they usually do not significantly impact the sentence’s meaning. If these types of words are not present, the module then masks auxiliaries, pronouns, and numbers, which can partly affect the sentence’s semantics. Finally, the module masks adjectives and verbs, which are critical for the sentence’s meaning. If there is only one non-noun word remaining or only nouns remain in the sentence, no further masking is performed. However, even with this priority order, some important words may still get masked, introducing noise into the training. Nevertheless, we empirically demonstrate that the language comprehension improvement from masking operations outweighs the negative effects of introducing noise (shown in Tab.~\ref{result_s}). In all datasets, the number of words exceeds 3, and through three masking operations, we find that the majority of the masked words are prepositions, conjunctions, and qualifiers. Therefore, in most cases, this operation will not affect the meaning of the sentence.
\par \noindent \textbf{Qualitative Component Analysis.} 
\textit{Qualitative analysis of MRCS.} 
Fig.~\ref{fig:FA_vis} illustrates the visualization of prediction results with or without the MFSD module on the robust VG datasets. It shows that the MFSD module enables the model to efficiently identify the presence or absence of targets described in the text on the image. The first row of the figure shows the false alarm data generated by the key nouns in the description being changed, the second row shows the false alarm data generated by the modification of key adjectives (e.g., color). The third line of the figure shows the spatial location relations in the description being modified and the fourth row of the figure shows the fine-grained features in the description being modified. Our MFSD module can effectively identify the false alarm data generated by all the above modification methods.
\textit{Qualitative analysis of MRCS.} Fig.~\ref{fig:kp_vis} presents the visual-linguistic feature map with or without MRCS module. We intuitively observe that the MRCS enables the feature map to attend more accurately to the target object's location, and generates a more precise foreground map. To avoid interactions from IMVF module, we conduct this experiment only with MRCS module and MFSD module. \textit{Qualitative analysis of IMVF.} Fig.~\ref{fig:TA} illustrates the visual-linguistic feature map with or without IMVF module. The figure indicates that the IMVF module reduces interference and allows the model to concentrate more on target by better understanding visual and textual information. To ensure fairness, we performed this experiment only with IMVF module and MFSD module.

\section{Conclusions}

Our work introduces the IR-VG framework, which comprises IMVF, MRCS, and MFSD. It outperforms existing approaches in terms of context features, fine-grained features, and localization accuracy while addressing robustness issues when faced with irrelevant or inaccurate reference expressions. Our experiments demonstrate the effectiveness of each module, achieving new SOTA performance.

\noindent \textbf{Limitation and future work.} Notably, IR-VG builds a new research direction for robust VG. Future work includes developing a more elegant framework to handle false alarms. In addition, we will explore the false-alarm problems with irrelevant expression for some foundation models (e.g. Grounding DINO~\cite{GDINO}).

{\small
\bibliographystyle{ieee_fullname}
\bibliography{egbib}
}

\end{document}